# Parallel Belief Revision

Daniel Hunter
Northrop Research and Technology Center
One Research Park
Palos Verdes Peninsula, CA. 90274

**Abstract**

This paper describes a formal system of belief revision developed by Wolfgang Spohn and shows that this system has a parallel implementation that can be derived from an influence diagram in a manner similar to that in which Bayesian networks are derived.

## 1  Introduction

This note describes a parallel architecture for a system of belief revision known as the "Spohn" system after its originator, Wolfgang Spohn [2]. First, a sketch of the Spohn system will be given. Next, a parallel implementation for the Spohn system analogous to Pearl's Bayesian Networks [1] will be described. It will be shown that when certain conditional independence conditions for propositional variables obtain, then (1) a network of marginal beliefs states (over subsets of propositional variables) suffices to reconstruct the joint belief state (over the set of all propositional variables) and (2) belief updating can be done in parallel in such a network.

## 2  Spohnian Belief Revision

Spohn's theory of belief revision concerns the revision of *deterministic* belief. Deterministic belief is *not* a matter of degree: one either believes, disbelieves, or does neither. A characteristic of deterministic belief is that for ideal reasoning agents, deterministic belief is closed under logical implication: whatever is logically implied by one's beliefs is also a belief. In contrast, for probabilistic belief, something implied by statements that are individually highly probable may be improbable.

Let $\Theta = \{s_1, s_2, ..., s_n\}$ be a set of pairwise mutually exclusive and jointly exhaustive states of the world. In Spohn's theory, a state of belief is captured by an *ordinal conditional function* (OCF), which is a function from $\Theta$ into ordinals that assigns zero to at least one member of $\Theta$. Although Spohn allows infinite ordinals, I will simplify matters by assuming an OCF's range to be a subset of the non-negative integers. Intuitively, an OCF is a grading of states of the world in terms of their degree of *implausibility*. That is, the greater the value of the OCF for a given state, the more implausible that state is.

A *proposition* is a statement whose truth-value depends upon which member of $\Theta$ is the true state of the world. Propositions are true or false in states. We follow the convention of identifying a proposition with the set of states in which it is true. Formally, then, a proposition is simply a subset of $\Theta$. An OCF $\kappa$ can be extended to consistent (i.e. nonempty) propositions by defining for each nonempty subset $A$ of $\Theta$:

$$\kappa(\mathbf{A}) = \min\{\kappa(s_i)|s_i \in \mathbf{A}\}.$$

An OCF $\kappa$ induces a *strength of belief* function $\beta$ over proper subsets $A$ of $\Theta$ as follows:

$$\beta(A) =_{df.} \begin{cases} -\kappa(A) & \text{if } \kappa(A) > 0 \\ \kappa(\neg A) & \text{otherwise} \end{cases}$$

It might seem from the above that the goal of analyzing deterministic belief has been forgotten and that the quite different notions of degrees of implausibility and strengths of belief have been introduced in its stead. However, deterministic belief is definable in this framework: A proposition $A$ is *(deterministically) believed* if and only if $\kappa^{-1}(0) \subseteq A$. In more intuitive terms, a proposition is believed if and only if it is true in all the most plausible worlds. If $A$ is a proper subset of $\Theta$ this is equivalent to saying that $A$ is believed if and only if $\beta(A) > 0$. Note that belief is closed under logical implication on this definition of belief.

Now we can say how beliefs are revised when new information is obtained. Suppose that proposition $P$ is learned with strength $\alpha$. Let $\kappa$ be the OCF before $P$ is learned. Then we define $\kappa'$, the OCF that results from learning $P$, by its value for each state $s$ as follows:

$$\kappa'(s) = \begin{cases} \kappa(s) - \kappa(P) & \text{if } s \in P \\ \kappa(s) - \kappa(\neg P) + \alpha & \text{otherwise} \end{cases}$$

As shown in [2], Spohnian belief revision has several nice properties:

1. It automatically restores consistency to beliefs when information inconsistent with previous beliefs is received.

2. It is commutative: learning $a$ and then learning $b$ results in the same OCF as first learning $b$ and then learning $a$, when $a$ and $b$ are independent.

3. It is easily reversible: if $\kappa_2$ results from $\kappa_1$ by learning proposition $a$ with strength $\alpha$, then $\kappa_1$ will result from $\kappa_2$ by learning proposition $\neg a$ with strength $\beta_1(\neg a)$, where $\beta_1$ is the strength of belief function corresponding to $\kappa_1$.

4. Notions of independence and conditional independence are definable and theorems analogous to those for probability theory are provable for these notions.

An example may aid understanding of Spohnian belief revision. Figure 1 shows three different OCFs for the six states corresponding to the rows of the table. The first OCF, for time $T_0$, gives one's beliefs about some unknown object.

|   |              |          | T0 | T1 | T2 |
|---|--------------|----------|----|----|----|
| 1 | PENGUIN      | FLYS     | 2  | 2  | 1  |
| 2 | PENGUIN      | NOT-FLYS | 1  | 1  | 0  |
| 3 | TYPICAL-BIRD | FLYS     | 0  | 0  | 1  |
| 4 | TYPICAL-BIRD | NOT-FLYS | 1  | 1  | 2  |
| 5 | NOT-BIRD     | FLYS     | 0  | 1  | 2  |
| 6 | NOT-BIRD     | NOT-FLYS | 0  | 1  | 2  |

Figure 1: Spohnian Belief Revision

At $T_1$, one learns with strength 1 that the object is a bird and one's beliefs change accordingly. Note that the proposition BIRD is identified with the set of states $\{1,2,3,4\}$. Then at $T_1$, one believes that the object flies. At $T_2$ one learns with strength 1 that the object is a penguin and the OCF is again revised. In the OCF for time $T_2$, one believes that the object does not fly.

## 3 Spohnian Networks

This section shows how a Spohnian network, like a Bayesian network, can be derived from an influence diagram.

For our purposes, we may take the nodes in the influence diagram of Figure 2 to be multi-valued variables. For example, node C might be a variable whose values are possible diagnoses for a certain patient. The values are assumed to be mutually exclusive and exhaustive.

In the following, I will use "a", "b", "c", "d", and "e" to denote arbitrary values of A, B, C, D, and E, respectively; more generally, a small letter will be used to denote an arbitrary value of a variable denoted by the corresponding capital letter. A set of variables will be denoted by a capital Greek letter and the corresponding small Greek letter will be used to denote a set of values of those variables.

An influence diagram is often given a causal interpretation in which a node is causally dependent upon all the nodes which have arrows pointing to it. But it can also be given an epistemic interpretation: it reveals certain conditional independencies between variables. Let $X$ and $Y$ be variables and $\Gamma$ a set of variables. Then we say that $X$ and $Y$ are independent given $\Gamma$ if all (undirected) paths between $X$ and $Y$ are separated by $\Gamma$. The notion of *separation* used here is a special one, defined as follows: We say that a path is *separated* by $\Gamma$ if and only if the path contains a pair of links meeting head-to-tail or tail-to-tail at a member of $\Gamma$. Links that meet head-to-head cannot separate a path. Thus in Figure 2, $D$ is separated from $A$ and $E$ by the set $\{B,C\}$, but $B$ is *not* separated from $C$ by $\{D\}$ or by any other set, since $D$ is the only node on the path from $B$ to $C$ and the links meeting at $D$ meet head-to-head. Hence $D$ is independent of all other variables given $B$ and $C$, but $B$ and $C$ themselves cannot be asserted to be independent given any subset of the remaining variables (including the empty

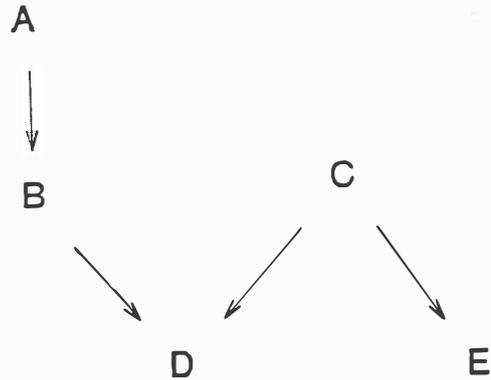

Figure 2: An Influence Diagram

set)[1].

A singly-connected (no more than one *undirected* path from a node to any other node) influence diagram can be turned into a computational structure for belief revision by associating with each node an OCF over that node and its parents. We call the resultant structure a *Spohnian network*. In the following, we will assume that we are always dealing with singly-connected influence diagrams.

A Bayesian network uses the conditional independencies of the corresponding influence diagram to break down a joint distribution into a function of a number of marginal distributions. We wish to show that a similar decomposition holds for Spohnian networks.

First, we need to define the relevant notion of conditional independence for the Spohn system. We begin with some notation. If $v_1, v_2, ..., v_k$ are values of the multi-valued variables $V_1, V_2, ..., V_k$, respectively, then $\kappa(v_1, v_2, ..., v_k)$ is the degree of implausibility of the proposition "$V_1$ has value $v_1$ and $V_2$ has value $v_2$ and ... and $V_k$ has value $v_k$." Let X, Y, Z be multi-valued variables and $\kappa$ an OCF. $\kappa(x|y)$ is defined to be $\kappa(x, y) - \kappa(y)$. Then we may define "X is independent of Y given Z with respect to $\kappa$" by:

For all values $x$ of X, $y$ of Y, and $z$ of Z, $\kappa(x|y, z) = \kappa(x|y)$.

If we interpret the independencies represented in an influence diagram as conditional independencies with respect to OCF's, then according to the influence diagram of Figure 2, the joint OCF over A, B, C, D, and E can be written as the following function of marginal OCF's:

$$\kappa(a, b, c, d, e) = \kappa(a, b) + \kappa(b, c, d) + \kappa(c, e) - \kappa(b) - \kappa(c).$$

The above formula shows that the Spohnian network derived from the influence diagram of Figure 2 correctly decomposes the joint OCF.

---

[1] Note that this definition of "separation" differs slightly from that in [1], which would count $B$ and $C$ independent.

# 4 Updating on a Single Piece of Uncertain Evidence

This section shows that updating on a single piece of uncertain evidence can be done in parallel in a Spohnian network.

Consider an arbitrary node $C$ in a Spohnian network and let N be the set $\{P_1, ..., P_r, C\}$, where $P_1, ..., P_r$ are all the parents of $C$. Then $C$ contains a marginal OCF over N. Let an update originate with variable $X$. First, we note that in the influence diagram there is a unique member of N through which all members are connected to $U$ (if they are connected at all). That is, there is a member $Y$ of N such that for any member $Z$ of N, if there is a path from $U$ to $Z$, then $Y$ lies along that path. This is because the influence diagram is singly-connected. Then we have the following theorem[2]:

**Theorem 1** *Let OCF $\kappa'$ result from OCF $\kappa$ by updating on variable $U$. Let $C$ be a variable with parents $P_1, ..., P_r$ and let $N = \{C, P_1, ..., P_r\}$. If $U$ is not connected to any member of $N$, then $\kappa'(c, p_1, ..., p_r) = \kappa(c, p_1, ..., p_r)$. Otherwise, let $Y$ be the unique member of $N$ through which all other members are connected to $U$ and let $\{Q_1, ..., Q_r\} = \{C, P_1, ..., P_r\} - \{Y\}$. Then $\kappa'(c, p_1, ..., p_r) = \kappa(q_1, ...q_r|y) + \kappa'(y)$.*

The above theorem shows that an update can be propagated through a Spohnian network in parallel. For if we know the update for variable $Y$, the update for the set $N$ can be computed. But the same theorem shows that the update for $Y$ depends only upon the update for $Y$'s immediate neighbor along the path from $U$ to $Y$ (if $Y = U$, we take this neighbor to be $Y$ itself). And the update for the neighbor depends only upon the update for its neighbor along the path from it to $U$ and so on. Since the influence diagram has no loops, these updates do not depend upon the updates for $Q_1, ..., Q_r$ and so $Y$ can be updated as soon as the update for its neighbor along the path from $U$ to $Y$ is computed. Hence an update can propagate through a Spohnian network in parallel. If, for example, in the network of Figure 2, an update originates at $C$, it can be propagated in parallel to each of the nodes to which it is connected.

An issue that arises at this point is whether or not updating preserves the relations of conditional independency. Although some conditional independencies may hold before updating but not after, the next theorem assures us that after updating, the parents of a variable continue to "screen off" that variable from other ancestral variables. Thus the decomposition into marginal OCF's will continue to hold after updating.

**Theorem 2** *Let $C$ be a variable and $\vec{P}$ a sequence of variables containing all the parents of $C$ in the influence diagram for OCF $\kappa$. If OCF $\kappa'$ arises from $\kappa$ by updating on variable $U$, then for any sequence $\vec{A}$ of ancestors of $C$, $\kappa'(c|\vec{p}, \vec{a}) = \kappa'(c|\vec{p})$.*

---

[2]Space does not permit proofs to be given here. Proofs of the theorems cited in this paper are available upon request.

# 5 Simultaneous Updating on Multiple Evidence Events

The previous section showed that a single piece of evidence may propagate through a Spohnian network in parallel. This section address the question of whether or not simultaneous multiple updates can propagate in parallel.

Unfortunately, if $X$ and $Y$ are dependent variables, then updating first on $X$ and then on $Y$ will not in general give the same result as updating in the reverse order. Thus simultaneous updating on dependent evidence events is problematic, since it is not clear what the answer should be.

One can approach the problem of simultaneous multiple updatings by first considering a special case in which the above problem does not arise. This is the case in which all the evidence events are learned with certainty. The strategy will be to prove that updating on several pieces of *certain* evidence can be done in parallel and then to argue that this case is not so special as it may appear.

To allow for learning with certainty in the Spohn system, I extend the notion of an OCF to allow $\infty$ to be assigned to states. $\infty$ is a number such that adding or subtracting any finite number from it leaves it unchanged. To learn a proposition with certainty, then, is to assign $\infty$ to all states incompatible with that proposition, indicating that all those other states are "infinitely implausible."

Another way of saying that a proposition is known with certainty is to say that its strength of belief is $\infty$. Hence $\kappa_{p,\infty}$ denotes the OCF that results from learning $p$ with certainty. From the updating rule, we see that for any $q$ compatible with $p$, $\kappa_{p,\infty}(q) = \kappa(q|p)$, while for any $r$ incompatible with $p$, $\kappa_{p,\infty}(r) = \infty$.

The next thing to note is that if proposition $p$ is learned with certainty, then it remains certain when any other *compatible* proposition is learned with any degree of strength.

Let $C$ be an arbitrary node in a Spohnian network and suppose that the new evidence is learned with certainty. Node $C$ will contain an OCF over the set of variables $N = \{C, P_1, ..., P_r\}$, where $P_1, ..., P_r$ are the parents of $C$. The evidence nodes may be partitioned into $r+1$ sequences: first, the sequence $\vec{D}$ of variables such that $C$ separates $\vec{D}$ from all the other variables in N; and the $r$ sequences $\vec{A}_i$ such that $P_i$ separates $\vec{A}_i$ from all other members of N. If a sequence $\vec{s}$ is empty, we stipulate that $\kappa(x|\vec{s}) = \kappa(x)$, for any value $x$ of any variable $X$.

Define $\Delta_c$ to be $\kappa(c|\vec{d}) - \kappa(c)$ and $\Delta_{p_i}$ to be $\kappa(p_i|\vec{a_i}) - \kappa(p_i)$ for $i = 1, ..., r$. Then it is easy to show that the final updated OCF over N is given by:

$$\kappa(c, p_1, ..., p_r) + \Delta_c + \sum_{i=1}^{r} \Delta_{p_i} -$$
$$\min\{\kappa(c^*, p_1^*, ..., p_r^*) + \Delta_{c^*} + \sum_{i=1}^{r} \Delta_{p_i^*} : c^* \in C, p_i \in P_i, i = 1, ..., r\}.$$

This result suggests the following algorithm for updating on multiple certain events. First, if $n_1, ..., n_s$ is any sequence of integers, positive or negative, define the *s-normalization* of $n_1, ..., n_s$ to be the sequence $n_1 - m, ..., n_s - m$, where $m$ is $\min\{n_1, ..., n_s\}$. An s-normalized sequence of integers will be a possible OCF since all its members will be non-negative and it will have at least one

175

zero member. The updating algorithm then says that where N is a node as described above, update each state $< c, p_1, ..., p_r >$ of N by adding to it the updates $\kappa(p_i|\vec{a_i}) - \kappa(p_i)$ and the update $\kappa(c|\vec{d}) - \kappa(c)$. After all updates are in, s-normalize the resultant vector of integers.

Two issues arise here. The first has to do with the term $\kappa(p_i|\vec{a_i})$. Since our updating scheme is asynchronous, not all the updates in $\vec{a_i}$ need arrive at $P_i$ at the same time. Do we have to wait until all these updates arrive at $P_i$ before passing along the updated degree of implausibility of $p_i$?

We don't. If the node for $P_i$ simply passes along the *change* in $p_i$'s implausibility after each update of $p_i$, then we will get the same result as if we had waited for all the updates of $p_i$. For suppose $b_1, ..., b_k$ are the updates in $\vec{a_i}$ in the order in which they arrive at $P_i$. Then the total update for each $p_i$ in $P_i$ will be

$$\kappa(p_i|b_1) - \kappa(p_i) + \sum_{j=2}^{k}(\kappa(p_i|b_1, ..., b_j) - \kappa(p_i|b_1, ..., b_{j-1})) = \Delta_{p_i}.$$

Hence if *changes* in degree of implausibility are passed along, then the result will be the same as if all the updates in $\vec{a_i}$ were assimilated by $P_i$ simultaneously. The same argument applies to $C$.

The second issue concerns the effect of updates coming through $P_j$ on the updating of $P_i$, for $j \neq i$. Let $\vec{b}$ be the updates from $\vec{a_i}$ received by $P_i$ up to some time and let $\vec{u}$ be all other updates received by $P_i$ up to that time. In the above reasoning, we ignored the effect of $\vec{u}$ on $P_i$, assuming that the update to $P_i$ would be $\kappa(p_i|\vec{b}) - \kappa(p_i)$, for each $p_i \in P_i$. Although this assumption is not technically correct, we can show that taking $\vec{u}$ into account does not affect the final update. Since only changes in degrees of implausibility are passed, the cumulative effect of the updates in $\vec{b}$ is to add to each $p_i$ the quantity $\kappa(p_i|\vec{u},\vec{b}) - \kappa(p_i|\vec{u})$. But $P_i$ separates $\vec{U}$ from $\vec{B}$, from which it easily follows that

$$\kappa(p_i|\vec{u},\vec{b}) - \kappa(p_i|\vec{u}) = \kappa(p_i|\vec{b}) - \kappa(p_i) + (\kappa(\vec{b}) + \kappa(\vec{u}) - \kappa(\vec{u},\vec{b}))$$

Hence for each $p_i$ in $P_i$ we add $\kappa(p_i|\vec{b}) - \kappa(p_i)$ plus some constant. And since the constant is the same for each $p_i$, it will have no effect on s-normalization. So the result of taking the updates $\vec{u}$ into account is the same as would obtain if only the updates $\vec{a_i}$ were used to update $P_i$.

# 6 Updating on Multiple Pieces of Uncertain Evidence

If simultaneous updates originate at variables $V_1, ..., V_n$, but these updates do not assign $\infty$ to specific values of these variables, there is a way of propagating these updates in parallel. Following the technique described in [1] for uncertain evidence in Bayesian networks, we add dummy nodes for binary variables $D_1, ..., D_n$ to the influence diagram with arrows pointing from $V_i$ to $D_i$. We label the values of $D_i$ by "$d_i$" and "$\bar{d_i}$." The Spohnian network corresponding to the augmented influence diagram will contain nodes with OCF's over the pairs

176

$\{V_i, D_i\}$. The trick is to choose the OCF's so that $\kappa(v_i|d_i)$ equals the original updated value of $v_i$, for each $v_i$ in $V_i$. Next, we set each value $\bar{d}_i$ to $\infty$. Then these updates on certain evidence are propagated as described in the previous section.

One must be aware, however, that this method will not in general produce a joint OCF in which the variables $V_1, ..., V_n$ have their original updates. This method seems to model a situation in which independent sources give updates for the $V_i$, one source for each variable, and the task is to combine the information from these sources. A different situation is one in which a single source stipulates the the $V_i$ are to have certain new marginal OCF's and the task is to find some revised joint OCF that satisfies these marginals. The difference between these two tasks must be kept in mind in deciding whether or not the updating method described in this section is appropriate.

## 7 Discussion

We have described Wolfgang Spohn's system of belief revision and shown that there is an efficient parallel implementation of that system. The Spohn system of belief revision has been implemented in Franz Lisp and a C version is being tested for use on the BBN Butterfly computer, a shared memory, parallel processor. An application of the Spohn system to fault diagnosis will be described in a future report.

In my opinion, Spohnian belief revision offers an elegant and novel approach to many of the problems in the area of uncertain reasoning. The example of Figure 1 illustrates how Spohnian belief revision easily handles the problem of non-monotonic reasoning. A related problem that arises in automatic planning – i.e., the Frame Problem or the problem of determining what changes and what stays the same when an action is performed – also becomes pleasantly tractable when cast in terms of Spohnian belief revision. Spohnian belief revision also has application to causal and counterfactual reasoning (in fact, Spohn was led to his system through a consideration of how causation might be defined within the realm of deterministic belief). Space does not permit elaboration on these applications, which will be explored in future work. They are mentioned here merely to give the reader a sense of the rich potential of Spohnian belief revision.

## References

[1] Pearl, J., "Fusion, Propagation, and Structuring in Bayesian Networks," *Artificial Intelligence*, 29, 1986, pp. 241-288.

[2] Spohn, W., "Ordinal Conditional Functions: A Dynamic Theory of Epistemic States," in *Causation in Decision, Belief Change, and Statistics, II*, ed. W.L. Harper and B. Skyrms, D. Reidel, 1988.